\documentclass[11pt]{article}

\usepackage{acl}

\usepackage{microtype}
\usepackage{graphicx}
\usepackage{booktabs}
\usepackage{tabularx}
\usepackage{array}
\usepackage{multirow}
\usepackage{amsmath}
\usepackage{amssymb}

\usepackage{fontspec}
\newfontfamily\ipafont{CharisSIL}[
  Extension=.ttf, UprightFont=*-Regular, BoldFont=*-Bold,
  ItalicFont=*-Italic, BoldItalicFont=*-BoldItalic]

\newcommand{\ipa}[1]{{\ipafont #1}}

\usepackage[title]{appendix}

\newcommand{\Ortho}{\textsc{Ortho}}
\newcommand{\PGoriely}{\textsc{P-Goriely}}
\newcommand{\PBreaks}{\textsc{P-breaks}}
\newcommand{\ProsodyNN}{\textsc{Prosody-NN}}

\title{Prosodic Structure as a Critical Component in Phonemic Language Modelling at BabyLM Scale}

\author{
    Bill K. S. Ku, Janet B. Pierrehumbert \\
    Department of Engineering Science \\
    University of Oxford \\  
    }

\begin{document}
\maketitle

\begin{abstract}
Human children learn language from continuous, prosodically structured speech
long before they can read, yet almost all language models train on orthographic
text. Building on previous phonemic BabyLM training \citep{goriely2024babble}, we ask whether supplying an estimate of the prosody infants exploit improves phonemic language modelling at BabyLM scale. We compare an orthographic baseline to three conditions forming a ladder of phonological richness. All conditions are trained as BERT MLMs at 10M and 100M words. Addressing the ecological validity of existing metrics in the small-data regime, we build a lexical decision task and two semantic organisation tasks.\footnote{The lexical decision and semantic evaluation sets will be distributed through \url{https://github.com/billkuai/prosodic_language_modelling_babylm} after the BabyLM Workshop at EMNLP 2026, in late Oct 2026.} Our new lexical decision task offers improved insights into the distinction between wordlikeness, token familiarity, and lexicality as factors. Triphone phonotactic control and pseudo-log-likelihood scoring shows that these scores reflect phonotactic familiarity rather than wordhood: every model sits near or below chance on an adversarial inverted condition, revealing prior results of this kind as partly artefactual. On our new semantic tasks, a prosodically annotated phoneme stream gives the strongest taxonomic organisation at both the 10M and 100M word scales. Phonemic models with boundary structure are competitive with the orthographic baseline. These findings suggest that prosodic structure is a key component of phonemic language modelling in the small-data regime most analogous to infant learning.
\end{abstract}

\section{Introduction}
\label{sec:intro}

Human children reach near-adult linguistic competence from a few tens of
millions of words of input, whereas state-of-the-art language models require
several orders of magnitude more data \citep{warstadt2023babylm,
fridman2024yannlecun}. The BabyLM Challenge \citep{warstadt2023babylm} reframes
this gap as a question of sample efficiency, capping training at
developmentally plausible budgets. A second and equally basic difference concerns the \emph{modality} of the signal. Children understand and produce language years before they learn to read: they acquire language from a continuous stream of speech in which
word boundaries are not marked and must be inferred. Almost all language
models, and most BabyLM submissions, instead train on orthographic text, in which word boundaries are marked with whitespace. 

Working in \emph{phonemic} space is a natural response \citep{goriely2024babble}. However, a plain phoneme stream discards prosody, a source of information that infants are known to use to bootstrap word segmentation and syntactic structure from the earliest stages of language acquisition. Here, we ask whether providing an estimate of the prosodic structure can enhance word learning from phonemic input at BabyLM scale. We compare an orthographic baseline (\Ortho) to three conditions of increasing phonological richness: a plain continuous-phoneme reconstruction of \citet{goriely2024babble} (\PGoriely); a condition that inserts intonation break tokens at major syntactic boundaries, as estimated from a dependency parse (\PBreaks); and one that further adds sentence-type boundary tones from a neural annotator trained on prosodically labelled speech (\ProsodyNN). Each condition is trained as a BERT masked language model
\citep{devlin2019bert} at two corpus scales, 10M and 100M words, yielding eight
models. 

Standard benchmarks use orthographic input. While conventions about the use of punctuation and whitespace in orthography reflect some prosodic features, such as the ends of utterances, they also reflect norms about which sequences are meaningful words. Some norms, such as the spacing around function words and elements of compounds, are arbitrary and language-specific -- clearly going beyond what human infants learn on a 10M budget, and in some respects beyond what children have learned after 100M words of language exposure. A focus on  learning from phonemic input thus aligns with the goals of the BabyLM project, in building models that draw inspiration from the efficiency of language learning by children.

Focusing on learning of the mental lexicon, we propose and evaluate three new tasks over child-relevant vocabulary.  A phonotactically controlled lexical decision task scored by pseudo-log-likelihood is similar to the BabySLM task in \citet{lavechin2022babyslm}. However, we redesign this task to make a stronger distinction between wordlikeness (having the phonological form of a word) and lexicality (having a meaning associated with the form).  Our semantic tasks, concept boundary discrimination and unsupervised category induction, are grounded in research on language development. 

Our contributions are threefold. In our redesign of the lexical decision task, we aim to support strong comparisons amongst wordlikeness, token familiarity, and lexicality. Pseudoword candidates are generated with a trigram Markov chain and scored in the phonemic domain.  A filter removes all candidates that are real words in either orthographic or phonemic form.  Then, pseudowords are individually selected to best match the phonotactic scores of the real words in the test. We also introduce an adversarial \emph{Inverted} test condition, in which each pseudoword has a higher phonotactic score than its real word competitor.  The results show that pseudo-log-likelihood scoring probes token familiarity more than lexical knowledge, and that previously reported phonemic-model results on the lexical decision task are partly artefactual. Second, we show that providing estimated prosodic markers  makes phonemic models competitive with, or better than,  the orthographic baseline on age-relevant semantic organisation. Taken together, these results indicate that, since the
infant's input is speech rather than text, careful phonemic representation
enriched with prosody is a key component in developmentally plausible language
modelling.

\section{Background and Related Work}
\label{sec:background}

\paragraph{Acquisition from continuous speech.}
Children reach near-adult linguistic competence from roughly 100 million
words of input by early adolescence \citep{warstadt2023babylm}. Before they learn to read, their input is speech: a continuous acoustic stream with no whitespace, no punctuation, and no explicit word boundaries. 
To build their mental lexicon, they segment words from fluent speech using distributional
regularities \citep{saffran1996statistical}, and exploit prosodic
structure to do so. The prosodic bootstrapping hypothesis
\citep{christophe2008bootstrapping, jusczyk1997discovery} holds that phrase
boundaries and boundary tones provide a partial parse of the signal before any
grammar is in place.  In addition, they carry information about some aspects of meaning. English yes/no questions typically carry a high (rising)
boundary tone while declaratives and wh-questions carry a low (falling) tone,
 a contrast that child-directed speech
exaggerates \citep{soderstrom2003prosodic}. Diphone statistics in combination with utterance boundaries are already rich enough to strongly support segmentation \citep{daland2011diphone}. These considerations motivate modelling language from a phoneme stream rather than from text.

\paragraph{Phonemic language modelling and BabyLM.}

Most BabyLM submissions train on orthographic text. \citet{goriely2024babble} is a notable exception.  They convert the BabyLM corpus to IPA with eSpeak-NG
\citep{espeakng}, remove word separators to form a continuous stream of
phonemes, and train a GPT-2 model, finding phonemic models to be fairly competitive with orthographic baselines. We extend this framework in three ways: we use the 
masked language model BERT throughout \citep{devlin2019bert}. Its bidirectional
objective is more data-efficient for the short utterances  that are typical in the training set. We make prosodic structure a primary experimental variable rather than discarding it; and we redesign the lexical decision task to remove a confound in prior evaluations.

\paragraph{Evaluating the mental lexicon.}
A word is an associative pair of a phonological form and a meaning
\citep{needle2022phonotactic}.  So, a lexical decision task should test whether a form is meaningful, not merely whether it is phonotactically word-like. Standard benchmarks serve this end imperfectly.   The speech-oriented BabySLM
\citep{lavechin2022babyslm} uses the pipeline in \citet{keuleers2010wuggy} and \citet{nguyen2020zero} to design  pseudowords for the forced-choice lexical decision task reported in \citet{goriely2024babble}. This pipeline does not control for triphone phonotactics, which are known to play an important role in wordlikeness judgments \citep{bailey2001determinants, hay2004speech, needle2022phonotactic}. 

We control for this factor, which intrinsically encompasses some aspects of syllable structure (though not all the aspects of syllable structure that BabySLM controls for). Overall, real words on average are expected to be phonotactically better than nonwords, because the encoding, memory, and production of lexical items is biased towards high phonotactic likelihood \citep{hay2004speech}. Through repeated learning and use, forms that include less likely sequences tend to be regularized towards more typical forms.   A matched-foil design that does not control for triphone likelihood thus retains a systematic asymmetry.  A post-hoc analysis of the BabySLM test set shows that the real word has better triphone phonotactics than the matched pseudoword in 70.3\% of the test pairs (see Appendix~\ref{app:wuggy-control}).  Thus, a model could score above chance simply by preferring the phonotactically better-formed item, without storing form-meaning associations.   The key distinction  between \emph{wordlikeness} and \emph{lexicality} is probed in our Inverted category (Section~\ref{sec:lexical}). 

The GLUE \citep{wang2018glue} and BLiMP
\citep{warstadt2020blimp} tasks are used in BabyLM papers to probe the learnability of syntax and semantics. After remapping the materials to a phonemic representation, \citet{goriely2024babble} reports accuracies in the range of 64\% to 78\% on BLiMP and GLUE for their various conditions and LMs, trained on 100M words. It is difficult to disentangle how much the errors result from weaknesses in the LMs versus intrinsic limitations of the training set size. Some patterns probed in these datasets are obviously not learned by humans from 10M words. For example, the word \textit{any} is learned only at age 4 or 5 \citep{tieu2016npi}, reflecting some 40M to 50M words of exposure. For a training budget of 64M tokens \citet{warstadt2020blimp} report accuracies of about 50\% for words like \textit{any}, and about 40\% for syntactic island constraints. Given this situation, we judged that the development of new simple semantic tasks with high face validity would best further our aim of evaluating how well prosodic information can support learning of lexicality. 

\section{Experimental Conditions}
\label{sec:conditions}
The orthographic text condition is included as a baseline. 
The phonological conditions form a ladder of increasing phonological richness, from the continuous phoneme stream as defined in \citet{goriely2024babble} to a phoneme stream annotated with phrase
boundaries and sentence-type boundary tones (Table~\ref{tab:conditions}).
All conditions are produced by a single preprocessing pipeline parameterised
per condition, so only the input representation differs across them. Every
condition is trained identically at both the 10M and 100M scales.

\subsection{Corpus}
\label{sec:corpus}
All conditions are trained on the BabyLM 2025 corpus \citep{charpentier2025babylmturns3papers},
which draws on six English sources (CHILDES, Project Gutenberg, OpenSubtitles,
Simple Wikipedia, the spoken BNC, and Switchboard). Roughly half of the tokens are transcripts of conversational or
child-directed speech, but the prosodic features were not transcribed, so our study estimates them (see Section~\ref{sec:phonemisation}). The full composition is given  in Appendix~\ref{app:annotation}.

\subsection{The Four Conditions}
\label{sec:ladder}
\Ortho{} is lowercased orthographic text and serves as the baseline.
\PGoriely{} reconstructs the approach of \citet{goriely2024babble}: the corpus
is transcribed to a continuous IPA stream with word separators and punctuation
removed, leaving only phonemes with an utterance-boundary token \texttt{[UTT]}.
\PBreaks{} adds two levels of the ToBI break-index hierarchy
\citep{beckman1997tobi}, an intermediate-phrase token \texttt{[B2]} and an
intonation-phrase token \texttt{[B4]}, derived from a syntactic parse rather
than punctuation; it is the first condition in which a structural token can
appear in a sentence with no surface punctuation. \ProsodyNN{} keeps these
break tokens and additionally marks sentence-final intonation with a high
(rising) boundary tone \texttt{[H\%]} or a low (falling) tone \texttt{[L\%]},
assigned by a neural annotator. Table~\ref{tab:examples} shows all four
conditions on three example sentences chosen to expose the two thresholds:
S3 triggers a \texttt{[B2]} at a subordinate-clause onset with no comma, and
S2 receives \texttt{[H\%]} as a yes/no question.

\begin{table}[t]
  \centering
  \small
  \setlength{\tabcolsep}{4pt}
  \begin{tabular}{@{}ll@{}}
    \toprule
    \textbf{Condition} & \textbf{Added structure} \\
    \midrule
    \Ortho     & orthographic text (baseline) \\
    \PGoriely  & continuous IPA; \texttt{[UTT]} at utterance ends \\
    \PBreaks   & \,$+$ \texttt{[B2]}, \texttt{[B4]} phrase boundaries \\
    \ProsodyNN & \,$+$ \texttt{[H\%]}, \texttt{[L\%]} boundary tones \\
    \bottomrule
  \end{tabular}
  \caption{The four conditions, organised as a baseline plus three phonemic conditions of increasing richness.  The phonemic conditions carry no whitespace.}
  \label{tab:conditions}
\end{table}

\begin{table}[t]
  \centering
  \small
  \setlength{\tabcolsep}{3pt}
  \begin{tabularx}{\columnwidth}{@{}l>{\raggedright\arraybackslash}X@{}}
    \toprule
    \textbf{Condition} & \textbf{Rendered output} \\
    \midrule
    \multicolumn{2}{@{}l}{\textit{S1: ``The dog sat on the mat.''}} \\
    \Ortho     & the dog sat on the mat . \\
    \PGoriely  & \ipa{ðədɒɡsætɒnðəmæt} \texttt{[UTT]} \\
    \PBreaks   & \ipa{ðədɒɡsætɒnðəmæt} \texttt{[B4]} \\
    \ProsodyNN & \ipa{ðədɒɡsætɒnðəmæt} \texttt{[B4]}\,\texttt{[L\%]} \\
    \midrule
    \multicolumn{2}{@{}l}{\textit{S2: ``Did she eat?''}} \\
    \Ortho     & did she eat ? \\
    \PGoriely  & \ipa{dɪdʃiːiːt} \texttt{[UTT]} \\
    \PBreaks   & \ipa{dɪdʃiːiːt} \texttt{[B4]} \\
    \ProsodyNN & \ipa{dɪdʃiːiːt} \texttt{[B4]}\,\texttt{[H\%]} \\
    \midrule
    \multicolumn{2}{@{}l}{\textit{S3: ``He left because it was late.''}} \\
    \Ortho     & he left because it was late . \\
    \PGoriely  & \ipa{hiːlɛftbɪkɒzɪtwɒzleɪt} \texttt{[UTT]} \\
    \PBreaks   & \ipa{hiːlɛft} \texttt{[B2]} \ipa{bɪkɒzɪtwɒzleɪt} \texttt{[B4]} \\
    \ProsodyNN & \ipa{hiːlɛft} \texttt{[B2]} \ipa{bɪkɒzɪtwɒzleɪt} \texttt{[B4]}\,\texttt{[L\%]} \\
    \bottomrule
  \end{tabularx}
  \caption{Three example sentences rendered under each condition. 
  %IPA is set in the body font; structural tokens in \texttt{monospace}. 
  S3 shows a \texttt{[B2]} inserted at a subordinate-clause onset with no comma; S2 shows
    the \texttt{[H\%]}/\texttt{[L\%]} sentence-type distinction.}
  \label{tab:examples}
\end{table}

\subsection{Phonemic Transcription and Prosodic Annotation}
\label{sec:phonemisation}
The phonemic conditions are transcribed with eSpeak-NG \citep{espeakng} through
the \texttt{phonemizer} library \citep{Bernard2021}, with the word separator set
to empty so the output is a continuous IPA stream with no explicit word
boundaries. Prosodic tokens are inserted by a two-pass scheme that phonemises
each word separately and interleaves boundary tokens at the recorded positions.
In \PBreaks{}, \texttt{[B2]} and \texttt{[B4]} are derived from a spaCy
dependency parse \citep{honnibal2020spacy}; in \ProsodyNN{}, the boundaries and
sentence-final tones (\texttt{[H\%]}/\texttt{[L\%]}) are predicted by a
\texttt{bert-base} sequence labeller trained on the ToBI-annotated Boston
University Radio News Corpus \citep{ostendorf1995burnc}, with the tone head
gated on held-out $F_1$ and a rule-based fallback. (For the annotator and gating thresholds, see Appendix~\ref{app:annotation}.)

\subsection{Tokenisation}
\label{sec:tokenisation}
A Byte-Pair Encoding tokeniser 
\citep{sennrich2016bpe} is trained from scratch for each condition, on its preprocessed corpus; the
character inventory and merge statistics of a prosodically annotated IPA stream
differ substantially from orthographic English. Vocabulary size is 4{,}000 at
10M and 8{,}000 at 100M. For the roughly 50-symbol IPA alphabet, BPE
yields primarily diphone and triphone units. All structural tokens
(\texttt{[B2]}, \texttt{[B4]}, \texttt{[H\%]}, \texttt{[L\%]}, \texttt{[UTT]},
and the reserved BERT tokens) are filtered from the corpus before BPE training
and inserted at fixed vocabulary ids afterwards. This prevents the algorithm from learning spurious merges inside token strings. Annotating the stream lengthens it: The annotated conditions carry roughly 10\% more tokens than \PGoriely{}, though every phonemic condition still consumes fewer tokens than \Ortho{}. Appendix~\ref{app:token-budget} reports the token counts and Appendix~\ref{app:tokenisation} the segmentation behaviour.

\subsection{Model Architecture and Training}
\label{sec:model}
A BERT-family masked language model 
\citep{devlin2019bert} is trained from scratch for each condition. BERT was selected due to its data-efficiency at this scale. The model size is matched to corpus scale: a BERT-small configuration
(${\approx}7.7$M parameters) at 10M and a BERT-base configuration (${\approx}92$M parameters) at 100M. All
conditions share the same training recipe; full architecture and
training hyperparameters are given in Appendix~\ref{app:model}. 

\section{Evaluation}
\label{sec:eval}

We evaluate all eight models on three purpose-built tasks, each defined in the
same representational space as training and restricted to infant-relevant
vocabulary: a phonotactically controlled lexical decision task
(Section~\ref{sec:lexical}) and two semantic tasks: concept boundary discrimination (Section~\ref{sec:cbd}) and unsupervised category induction (Section~\ref{sec:sci}). All tasks read from the
frozen trained models without any fine-tuning.

\subsection{Lexical Decision}
\label{sec:lexical}

The lexical decision task asks whether a model prefers a real word to a
phonotactically matched pseudoword (except for \emph{Inverted} category pairs). Because the two items in a pair are matched on phonotactic plausibility (or, by deliberate design, favours the pseudoword for the \emph{Inverted} category) under the model's own training distribution, above-chance accuracy is evidence of stored lexical knowledge rather than
phonotactic filtering alone.

\paragraph{Phonotactic scoring.}
Each condition has its own trigram ($n{=}3$) phonotactic model, estimated from that model's training dictionary. Following \citet{needle2022phonotactic}, counts are type-based:
each word type contributes once regardless of corpus frequency, so the model
captures phonotactic \emph{licitness} rather than lexical \emph{frequency}. The
phonotactic score of a word $w$ with phone sequence $\phi_1 \dots \phi_L$ is the
log-probability of its padded form
$\Phi(w) = \langle s\rangle\,\langle s\rangle\,\phi_1 \dots \phi_L\,\langle/s\rangle$
under a Witten-Bell smoothed model \citep{witten1991}:
\begin{equation}
  S(w) = \sum_{i=3}^{L+3} \log
  P_{\mathrm{WB}}\!\left(\Phi_i \mid \Phi_{i-2}, \Phi_{i-1}\right).
\end{equation}
Witten-Bell smoothing interpolates each context with its lower-order backoff so
that unseen trigrams in the test set still receive probability mass. Writing $h$ for a context,
$c(h,\phi)$ for the count of $\phi$ following $h$,
$c(h) = \sum_{\phi'} c(h,\phi')$, and
$t(h) = |\{\phi' : c(h,\phi') > 0\}|$ for the number of distinct continuations
of $h$,
\begin{equation}
  P_{\mathrm{WB}}(\phi \mid h) =
  \begin{cases}
    \frac{c(h,\phi)}{c(h) + t(h)} & c(h,\phi) > 0, \\[1.4ex]
    \lambda(h)\, P_{\mathrm{WB}}(\phi \mid h') & c(h,\phi) = 0,
  \end{cases}
\end{equation}
where $h'$ drops the leftmost symbol of $h$ and the recursion bottoms out at a
uniform unigram floor. The backoff weight
$\lambda(h) = t(h)/\big(c(h)+t(h)\big)$ grows with the diversity of observed
continuations $t(h)$, allocating more probability to novel phones in contexts
that have shown lexical variety. For \Ortho{} the scorer operates on characters.

\paragraph{Pseudoword generation.}
For practical reasons, pseudowords are first generated in the orthographic space. Then they are mapped into the 
phonemic space using eSpeak-NG, and scored using the triphone Markov model described above. Pairing is carried out once, in the phonemic space of \PGoriely{}, to ensure every condition is tested on identical items. This means that the pseudowords are created to match the real-word phonotactic distribution by construction and lets
us control their score directly, which the Inverted category requires. Given a
context $\mathrm{ctx} = (\phi_{i-2}, \phi_{i-1})$, the next character is sampled with
a temperature $T$:
\begin{equation}
  P_T(\phi_i \mid \mathrm{ctx}) \;\propto\; c(\mathrm{ctx}, \phi_i)^{1/T},
\end{equation}
generating autoregressively from $\langle s\rangle\langle s\rangle$ until
$\langle/s\rangle$ is emitted. Each candidate is then phonemised with eSpeak-NG
and scored with the model above, and it is that phonemic score, not the
proposal, which decides what enters the test set. Lowering the temperature to
$T{=}0.3$ concentrates proposals on frequent transitions, yielding a pool dense
enough in high-scoring forms to supply the Inverted category. The full
procedure is given in Appendix~\ref{app:wuggy-control}.

Items are pairs of a real word and a pseudoword. 
Two filters guarantee that the items are valid. \emph{Filter A} admits a real
word only if it is a genuine English word, that is, attested in a reference word list after lemmatisation; this removes corpus artefacts,  such as \textit{https} or repeated-character strings that survive a purely alphabetic filter and would
otherwise be labelled real words. \emph{Filter B} admits a pseudoword only if
neither its orthographic form nor its eSpeak IPA form (raw, and with stress and length marks stripped) occurs in the training set, so that no pseudoword is
in fact a word the model has seen, whether spelled differently from a real
homophone or transcribed identically to one. Rejected candidates are resampled
so category sizes are held fixed. This phonemic screen is essential: without it, a large share of candidate pseudowords coincide with real words in IPA space and are
wrongly counted as model errors. Contamination statistics are reported in
Appendix~\ref{app:lexical}.
For the \Ortho{} condition, the real word is paired with the original orthographic form of the pseudoword, from which the phonemic form was generated. (Using the same test items across all conditions means that the orthotactic likelihood is not fully controlled). 

\paragraph{Pair categories.}
Real words fall into three categories. \emph{LexicalMemory} words are a single
token in the condition's BPE vocabulary, so the model can score the whole word
in one step. \emph{Composition} words span multiple tokens, requiring the model
to compose a representation from sub-word units. \emph{Inverted} pairs are
adversarial: the real word is drawn from the bottom of the phonotactic-score
distribution and the pseudoword from the top, so that
$S(w_{\mathrm{fake}}) > S(w_{\mathrm{real}})$ by construction. A model relying
only on phonotactics will therefore prefer the pseudoword and score at or below
chance on Inverted; only stored lexical knowledge that overrides phonotactic
expectation can raise accuracy above chance. Standard, non-inverted pairs are
matched to within $2.0$ log-probability units. The evaluation set is built once
on orthographic words and derived for the phonemic conditions by phonemising the
same pairs, so every model is tested on identical word content; it contains
$4{,}129$ pairs, of which $2{,}000$ are Inverted, with the split between
LexicalMemory and Composition depending on each condition's vocabulary.

\paragraph{Metric.}
Each model is scored by pseudo-log-likelihood (PLL; \citealp{salazar2020pll}),
the natural scoring rule for a masked language model. For a word tokenised as
$\mathbf{w} = (w_1, \dots, w_N)$,
\begin{equation}
  \mathrm{PLL}(\mathbf{w}) = \sum_{i=1}^{N}
  \log P_\theta\!\left(w_i \mid \mathbf{w}_{\setminus i}\right),
\end{equation}
the sum of masked-token log-probabilities across the word. A pair is judged
correct when $\mathrm{PLL}(w_{\mathrm{real}}) > \mathrm{PLL}(w_{\mathrm{fake}})$.
Accuracy is reported separately for each category; chance is $0.50$.

\subsection{Semantic Tasks}

Both semantic tasks operate on contextualised embeddings of 68 early-vocabulary
words drawn from the MacArthur-Bates Communicative Development Inventories
\citep{fenson1993cdi}, spanning six categories (animals, food, body parts,
clothing, vehicles, and household objects); the full list is given in
Appendix~\ref{app:semantic}. For each target word we collect up to 30 short
excerpts where it occurs from the CHILDES set
\citep{10.1162/coli.2000.26.4.657}.  render each excerpt into the surface form for the condition, with the frozen model. We use the \emph{target-anchored} representation: the per-excerpt vector is the mean of the final-layer hidden states at the subword positions of the target word, and the word vector is the mean over all excerpts. Writing $T_e$ for the target's subword positions in excerpt
$e \in E_w$ with hidden states $\mathbf{h}^{(e)}_t$,
\begin{equation}
  \mathbf{e}_w = \frac{1}{|E_w|} \sum_{e \in E_w}
  \frac{1}{|T_e|} \sum_{t \in T_e} \mathbf{h}^{(e)}_t .
\end{equation}
Anchoring on the target word, rather than mean-pooling the whole excerpt,
isolates the word's representation from the sentence topic and is what makes the
measure word-specific.

\subsubsection{Concept boundary discrimination (CBD)}
\label{sec:cbd}
{\bf Concept boundary discrimination (CBD)} 
%is an excerpt-level 
is a 3-way ranking task. 
For each target word, a probe presents three alternatives: a
\emph{same-category} word, a \emph{thematic foil} from a different category but
linked to the target by a non-taxonomic relation, and an
\emph{unrelated foil}. Two words are defined as thematically related when they stand in a
script or event association, a production or source relation, a containment or
part-whole relation, a spatial co-occurrence, or a functional relation, and not through shared category membership \citep{lin2001thematic, imai1994young}.
Each of the 68 targets is assigned a single thematic foil by a hand-curated
lookup; because the foil must itself be a CDI word to have CHILDES excerpts,
roughly 30 targets yield a valid foil, producing 300 probes at ten draws per
word. Table~\ref{tab:foils} gives representative foil pairs across the five
relation families.

\begin{table}[t]
  \centering
  \small
  \setlength{\tabcolsep}{5pt}
  \begin{tabular}{@{}lllc@{}}
    \toprule
    \textbf{Target} & \textbf{Thematic foil} & \textbf{Relation} & \textbf{CDI} \\
    \midrule
    \multicolumn{4}{@{}l}{\textit{Script / event}} \\
    cat   & milk   & typical food & \checkmark \\
    fish  & water  & habitat      & \checkmark \\
    dog   & bone   & typical food & \textdagger \\
    \midrule
    \multicolumn{4}{@{}l}{\textit{Production / source}} \\
    cow   & milk   & product      & \checkmark \\
    egg   & bird   & source       & \checkmark \\
    \midrule
    \multicolumn{4}{@{}l}{\textit{Containment / part-whole}} \\
    milk  & bottle & container    & \checkmark \\
    shoe  & foot   & worn on      & \checkmark \\
    pocket& hand   & contains     & \checkmark \\
    \midrule
    \multicolumn{4}{@{}l}{\textit{Spatial / locational}} \\
    boat  & water  & location     & \checkmark \\
    car   & road   & location     & \textdagger \\
    \midrule
    \multicolumn{4}{@{}l}{\textit{Functional / instrumental}} \\
    eye   & look   & function     & \textdagger \\
    cheese& mouse  & typical eater& \checkmark \\
    \bottomrule
  \end{tabular}
  \caption{Representative thematic foil pairs by relation family. Each foil is a
    different-category word tied to the target by a non-taxonomic relation and,
    to have CHILDES excerpts, must itself be a CDI target. Entries marked
    \textdagger{} have no CDI-resident foil and contribute no probes.}
  \label{tab:foils}
\end{table}

The model output is the cosine similarity of each of the three alternatives to the probe word. The measure of interest is the ranking of these three alternatives. The cosine similarity is defined by the following equation (with maximum similarity having the value $1.0$ ).

\begin{equation}
  s_{j \in \{\mathrm{same},\, \mathrm{them},\, \mathrm{unrel}\}} = \cos\!\left(\mathbf{e}_{\mathrm{tgt}},\, \mathbf{e}_j\right),
\end{equation}

We report three accuracies: three-way (the same-category word preferred over
both foils, chance $\tfrac{1}{3}$), taxonomic-versus-thematic,  and
taxonomic-versus-unrelated (the same-category word preferred over the thematic or
unrelated foil respectively, chance $\tfrac{1}{2}$). Taxonomic-versus-thematic is the primary metric: it isolates taxonomic organisation from thematic association, a distinction distributional models often fail to make. A category neighbour (e.g. 
  \textit{cat} for \textit{dog}) should be preferred to a merely thematic associate (e.g. \textit{bone}). 
Writing $s_{\mathrm{same}}$, $s_{\mathrm{them}}$ and $s_{\mathrm{unrel}}$ for the
cosine similarity between $\mathbf{e}_{\mathrm{tgt}}$ and the same-category word,
the thematic foil and the unrelated foil respectively, a probe is scored correct when:
\begin{equation}
  \begin{aligned}
    \text{3-way:}\quad    & s_{\mathrm{same}} > \max\!\left(s_{\mathrm{them}},\, s_{\mathrm{unrel}}\right), \\
    \text{Tv-them:}\quad  & s_{\mathrm{same}} > s_{\mathrm{them}}, \\
    \text{Tv-unrel:}\quad & s_{\mathrm{same}} > s_{\mathrm{unrel}}.
  \end{aligned}
\end{equation}
 We also report the mean taxonomic advantage: the average margin by which the
same-category similarity exceeds the thematic-foil similarity,
\begin{equation}
  \mathrm{adv} = \frac{1}{N} \sum_{n=1}^{N}
  \Big[ \cos\!\left(\mathbf{e}_{\mathrm{tgt}}, \mathbf{e}_{\mathrm{same}}\right)
      - \cos\!\left(\mathbf{e}_{\mathrm{tgt}}, \mathbf{e}_{\mathrm{them}}\right) \Big].
\end{equation}

\subsubsection{Semantic Category Induction (SCI)}
\label{sec:sci}

{\bf Semantic category induction (SCI)} tests whether word vectors, clustered
without supervision, recover the six ground-truth categories. The aggregated
vectors $\mathbf{e}_w$ of all 68 words are clustered into $K = 6$ groups with
both $K$-means (ten random initialisations) and agglomerative clustering under
Ward's linkage \citep{ward1963hierarchical}. Cluster quality is measured against
the category labels by the Adjusted Rand Index
\citep[ARI;][]{hubert1985comparing} and Normalised Mutual Information (NMI),
with $K$-means ARI as the primary metric. The ARI corrects the Rand Index (RI)
for chance agreement,
\begin{equation}
  \mathrm{ARI} = \frac{\mathrm{RI} - \mathbb{E}[\mathrm{RI}]}
                      {\max(\mathrm{RI}) - \mathbb{E}[\mathrm{RI}]},
\end{equation}
so that $\mathrm{ARI} = 0$ corresponds to random labelling and
$\mathrm{ARI} = 1$ to perfect categorisation. SCI is evaluated on the words
observed in training. It tests only how the model has organised the vocabulary it
has actually seen.

\section{Results}
\label{sec:results}

We report results for the three tasks, at both the 10M and 100M scales. All figures are read from the frozen models.

\subsection{Lexical Decision}
\label{sec:results-lexical}

Table~\ref{tab:pll-results} reports PLL accuracy by pair category. On
LexicalMemory, every condition is near ceiling (0.930 to 0.991). For Composition, the results range from 0.768 to 0.819; the phonemic conditions perform similarly to \Ortho{}, with \PGoriely{} highest at 10M and \Ortho{} highest at 100M.
On the Inverted category,  performance for every condition is poor.  Values sit close to the 0.50 chance level at 10M (0.479 to 0.506); at 100M \Ortho{} remains at chance (0.496) while the three phonemic conditions fall below it, to between 0.396 and 0.444.

\begin{table}[t]
  \centering
  \small
  \setlength{\tabcolsep}{5pt}
  \begin{tabular}{@{}lccc@{}}
    \toprule
    \textbf{Condition} & \textbf{LexMem} & \textbf{Comp} & \textbf{Inv.} \\
    \midrule
    \multicolumn{4}{@{}l}{\textit{10M}} \\
    \Ortho     & \textbf{0.991} & 0.768          & 0.499 \\
    \PGoriely  & 0.942          & \textbf{0.819} & \textbf{0.506} \\
    \PBreaks   & 0.948          & 0.799          & 0.489 \\
    \ProsodyNN & 0.930          & 0.800          & 0.479 \\
    \midrule
    \multicolumn{4}{@{}l}{\textit{100M}} \\
    \Ortho     & 0.980          & \textbf{0.813} & \textbf{0.496} \\
    \PGoriely  & \textbf{0.988} & 0.801          & 0.396 \\
    \PBreaks   & 0.982          & 0.808          & 0.444 \\
    \ProsodyNN & 0.979          & 0.807          & 0.421 \\
    \bottomrule
  \end{tabular}
  \caption{PLL accuracy on the lexical decision task ($n = 4{,}129$ pairs) by
    pair category. Chance is $0.50$. Bold marks the best condition in each
    column within a scale.}
  \label{tab:pll-results}
\end{table}

\subsection{Semantic Tasks}
\label{sec:results-cbd}

\begin{figure}[t]
  \centering
  \includegraphics[width=\columnwidth]{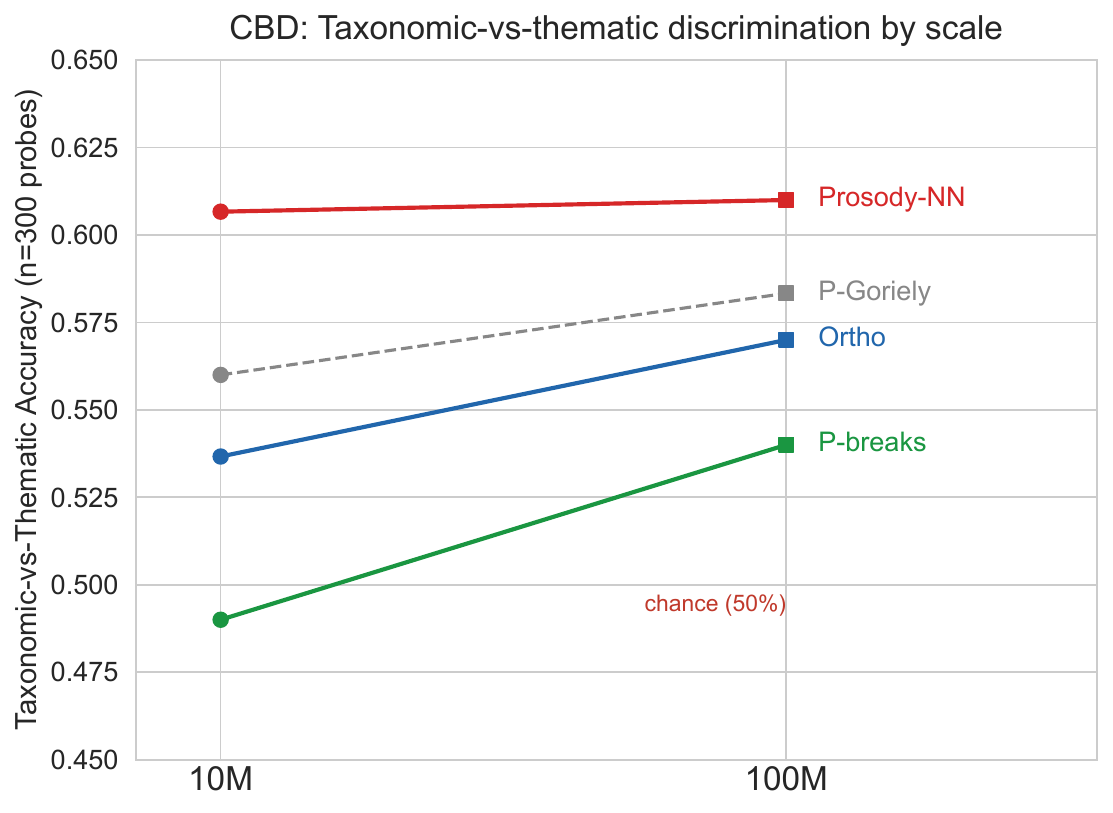}
  \caption{Taxonomic-versus-thematic accuracy at 10M and 100M for the four
    conditions. The dashed line marks chance ($0.50$).}
  \label{fig:cbd-scale}
\end{figure}

Table~\ref{tab:cbd-results} reports Concept Boundary Discrimination accuracy over 300 probes. On the primary taxonomic-versus-thematic metric, \ProsodyNN{} is highest at both scales, and no other condition exceeds 0.6; \ProsodyNN{}
also has the largest mean taxonomic advantage at both scales. \Ortho{} leads the easier taxonomic-versus-unrelated comparison at both scales. \PBreaks{} is lowest on taxonomic-versus-thematic at 10M, at 0.490. Figure~\ref{fig:cbd-scale} plots the taxonomic-versus-thematic accuracy across scales.

\begin{table}[t]
  \centering
  \small
  \setlength{\tabcolsep}{5pt}
  \begin{tabular}{@{}lcccc@{}}
    \toprule
    \textbf{Condition} & \textbf{3-way} & \textbf{Tv-them} & \textbf{Tv-unrel} & \textbf{Adv} \\
    \midrule
    \multicolumn{5}{@{}l}{\textit{10M}} \\
    \Ortho     & 0.463          & 0.537          & \textbf{0.693} & $+0.019$ \\
    \PGoriely  & 0.420          & 0.560          & 0.613          & $+0.028$ \\
    \PBreaks   & 0.400          & 0.490          & 0.600          & $+0.009$ \\
    \ProsodyNN & \textbf{0.483} & \textbf{0.607} & 0.670          & $\mathbf{+0.040}$ \\
    \midrule
    \multicolumn{5}{@{}l}{\textit{100M}} \\
    \Ortho     & \textbf{0.533} & 0.570          & \textbf{0.860} & $+0.019$ \\
    \PGoriely  & 0.477          & 0.583          & 0.737          & $+0.022$ \\
    \PBreaks   & 0.443          & 0.540          & 0.703          & $+0.013$ \\
    \ProsodyNN & \textbf{0.533} & \textbf{0.610} & 0.747          & $\mathbf{+0.027}$ \\
    \bottomrule
  \end{tabular}
  \caption{CBD accuracy ($n = 300$ probes; $\mathrm{SE} \approx 0.028$ on
    Tv-them). \emph{Tv-them}: taxonomic vs.\ thematic (primary);
    \emph{Tv-unrel}: taxonomic vs.\ unrelated; \emph{Adv}: mean taxonomic
    advantage. Chance is $\tfrac{1}{3}$ for 3-way and $0.50$ for the pairwise
    columns. Bold marks the best condition in each column within a scale.}
  \label{tab:cbd-results}
\end{table}

\begin{figure}[t]
  \centering
  \includegraphics[width=\columnwidth]{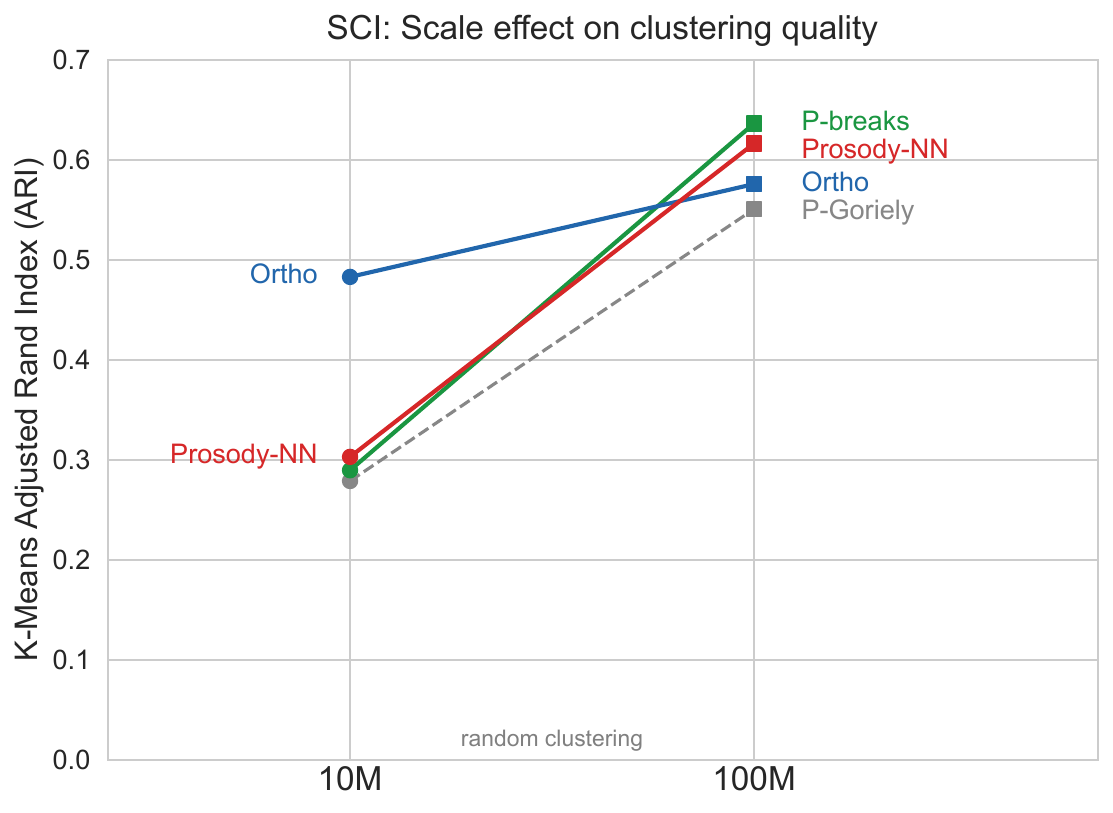}
  \caption{$K$-means ARI at 10M and 100M for the four conditions. \PBreaks{} and
    \ProsodyNN{} cross above \Ortho{} at 100M; the dashed line marks ARI $= 0$.}
  \label{fig:sci-scale}
\end{figure}

For Semantic Category Induction, Table~\ref{tab:sci-results} reports cluster recovery over the 68 words. At 10M,
\Ortho{} leads on all three measures (K-ARI 0.483), with the phonemic conditions
between 0.279 and 0.303. At 100M the ordering changes: \PBreaks{} (K-ARI 0.637)
and \ProsodyNN{} (0.617) exceed \Ortho{} (0.576) on both $k$-means measures,
while \Ortho{} leads on agglomerative ARI (0.717). Figure~\ref{fig:sci-scale}
shows the 10M-to-100M trajectories, in which \PBreaks{} and \ProsodyNN{} cross
above \Ortho{}. The detailed embedding geometry could be found in Appendix~\ref{app:umap}.

\begin{table}[t]
  \centering
  \small
  \setlength{\tabcolsep}{6pt}
  \begin{tabular}{@{}lccc@{}}
    \toprule
    \textbf{Condition} & \textbf{K-ARI} & \textbf{K-NMI} & \textbf{A-ARI} \\
    \midrule
    \multicolumn{4}{@{}l}{\textit{10M}} \\
    \Ortho     & \textbf{0.483} & \textbf{0.601} & \textbf{0.337} \\
    \PGoriely  & 0.279          & 0.439          & 0.257 \\
    \PBreaks   & 0.290          & 0.417          & 0.260 \\
    \ProsodyNN & 0.303          & 0.454          & 0.259 \\
    \midrule
    \multicolumn{4}{@{}l}{\textit{100M}} \\
    \Ortho     & 0.576          & 0.681          & \textbf{0.717} \\
    \PGoriely  & 0.551          & 0.666          & 0.503 \\
    \PBreaks   & \textbf{0.637} & \textbf{0.750} & 0.530 \\
    \ProsodyNN & 0.617          & 0.732          & 0.610 \\
    \bottomrule
  \end{tabular}
  \caption{SCI cluster-recovery scores ($n = 68$ words, $k = 6$). \emph{K-ARI},
    \emph{K-NMI}: $k$-means ARI and NMI (K-ARI primary); \emph{A-ARI}:
    agglomerative ARI. Bold marks the best condition in each column within a
    scale.}
  \label{tab:sci-results}
\end{table}

\section{Discussion}
\label{sec:discussion}

\subsection{Lexical Decision}
\label{sec:disc-lexical}

Pseudo-log-likelihood is strongly influenced by phonotactic familiarity. The near-ceiling accuracy for  LexicalMemory reflects only the fact that a single-token real word
receives strong probability under virtually any model. The accuracy for Composition is lower, but still above chance,  because the real word's multi-token sequence is partially learned while the pseudoword's multi-token sequence is unseen. The Inverted category is decisive: every condition sits near or below chance. This means that the task does not incisively probe lexicality; high wordlikeness can cancel out the partial ability to learn meaningfulness, that was seen in the Composition category. At 100M the phonemic conditions fall below chance while \Ortho{} stays at chance: the larger model has learned its input's phonotactics more sharply and scale deepens PLL's phonotactic dependence rather than
relieving it. These findings suggest that the high lexical accuracy in \citet{lavechin2022babyslm} for the PLL lexical decision task is partially an artefact of the BabySLM stimulus design. 

\subsection{Semantic Organisation}
\label{sec:disc-semantic}
%On both semantic tasks (Sections~\ref{sec:cbd} and~\ref{sec:sci}) the intended comparison holds. 
For the CBD task, \ProsodyNN{} beats the orthographic and plain-phonemic baselines on taxonomic-versus-thematic discrimination at both scales
(Figure~\ref{fig:cbd-scale}). For the SCI task, it rises above
\Ortho{} at 100M (Figure~\ref{fig:sci-scale}). 
\ProsodyNN{}'s performance indicates that its boundary and tone tokens reshape the embedding space toward categorical rather than merely thematic similarity. \Ortho{} keeps
the lead only on the easier taxonomic-versus-unrelated comparison, where co-occurrence alone separates a category neighbour from an unrelated word.

The behaviour of \PBreaks{} is striking.  It is the weakest
condition on taxonomic-versus-thematic, yet on cluster quality it matches or exceeds \ProsodyNN{} and rises above \Ortho{} at 100M, where it posts the highest $k$-means ARI. Break tokens therefore appear to supply a segmentation scaffold that
yields well-separated clusters, but on some basis other than taxonomy.  The categorical sharpening that CBD rewards appears to come from the sentence-type tone annotation that distinguishes \ProsodyNN{} from \PBreaks{}, not from the break structure alone.

A second conspicuous pattern is the slope of the scale effect. Every condition improves from 10M to 100M. However, the phonemic conditions climb far more steeply than \Ortho{} (Figure~\ref{fig:sci-scale}), converting a ten-fold increase in data into categorical organisation  more efficiently than \Ortho{} does. We read this as a difference in what each representation still has to gain. Whitespace gives \Ortho{} its word units for free, so its category structure rests on the co-occurrence with other words. By 10M, it has obtained almost as much advantage from the whitespace as is possible. In contrast, the phonemic conditions must induce the word boundaries, and prosodic tokens mark the phrase edges that shape which co-occurrences count. By 100M words, it appears that the phonemically-induced word boundaries may actually be better than those marked by whitespace in \Ortho{}, overcoming some of the idiosyncracies of English spelling and punctuation. The margins on any single probe set are small (the binomial standard error at $n = 300$ is about $0.028$), but the pattern is consistent across both tasks, both scales, and the more robust taxonomic-advantage measure; the tighter, isotropic clusters of the boundary-annotated conditions are visible in Appendix~\ref{app:umap}.

\subsection{Conclusion}
\label{sec:conclusion}

\citet{goriely2024babble} established that training on continuous phonemic streams is a strong competitor to orthographic text at BabyLM scale. Our work
extends that framework with explicit prosodic structure, a phonotactically controlled evaluation, and a comparison across two scales over four conditions. Two findings stand out. First, our redesigned lexical decision task shows that
pseudo-log-likelihood scoring is strongly influenced by phonotactic likelihood rather than probing lexical knowledge alone: every model sits around or below chance on the adversarial Inverted category, and accuracy moves with the phonotactic balance rather than with lexicality alone. Second, on the semantic tasks, prosodic annotation makes phonemic models competitive with, or in some cases better than, the orthographic baseline. 

Together these results indicate that, in the data-limited regime most analogous to infant learning, the structure and naturalness of the input representation matters. Orthographic text is not a neutral default but a cognitive artefact
whose statistics a model can exploit. Because the child's input is speech and not text, a carefully constructed phonemic representation enriched with prosody is a key component of developmentally plausible language models, and one that future BabyLM work can build on.

\clearpage

\section*{Limitations}

An important limitation of this study is that the prosodic
annotation is not observed from speech. Instead, it is estimated from text.  The boundary and tone tokens
are estimated using a dependency parser and a BERT sequence labeller, which was trained on the Boston University Radio
News Corpus.  Further, the  read broadcast-news register of this corpus differs from the conversational and child-directed speech that dominates the BabyLM corpus. Our annotations are therefore an approximation of the prosody an infant hears, and the gains we report are plausibly a lower bound on what genuine speech-based training would yield. Phonemisation adds a further approximation: eSpeak-NG applies a single American-English accent and rule-based grapheme-to-phoneme conversion, which introduces systematic transcription errors shared across all phonemic conditions.

The study used BPE encoding of the input, as this encoding is used in the baselines to which we make comparisons. When the white space marking word boundaries is removed, BPE encoding of an utterance can result in single tokens that bridge a (de facto) word boundary; see Appendix~\ref{app:tokenisation} for further details. We do not consider alternative character-based tokenisation schemes. These have provided promising results in some BabyLM studies~\citep{goriely2024babble, bunzeck-zarriess-2025-subword, bunzeck-etal-2025-small}. Exploration of character-based models for our experimental conditions and tasks is an important direction for future research. 

The evaluations are rather narrow. The semantic tasks use 68 high-frequency CDI words in six broad categories, with hand-curated thematic foils yielding 300 probes, so the margins on any single metric are small and the categories may miss finer-grained structure; low-frequency and out-of-vocabulary
generalisation is untested. All models are trained with a single random seed, so we report no variance estimates, and because the two scales differ in both corpus size and model capacity, scale effects conflate the two and support only
within-scale condition rankings. Finally, our lexical decision analysis relies on pseudo-log-likelihood alone; a discriminative probe of the same representations, and a full account of why \PBreaks{} forms high-quality clusters on a non-taxonomic basis, are left to future work, as is the natural step of replacing text-derived annotation with acoustic prosody from speech or self-supervised speech units.

\section*{Acknowledgments}

We are grateful to Marvin Lavechin for his prompt and helpful responses to our questions about the construction of the BabySLM dataset. We also appreciate the helpful input from members of the Pierrehumbert Language Modelling Group over the course of the project.

\bibliography{references}

@article{bailey2001determinants,
  title={Determinants of wordlikeness: Phonotactics or lexical neighborhoods?},
  author={Bailey, Todd M and Hahn, Ulrike},
  journal={Journal of Memory and Language},
  volume={44},
  number={4},
  pages={568--591},
  year={2001},
  publisher={Elsevier}
}

@article{nguyen2020zero,
  title={The zero resource speech benchmark 2021: Metrics and baselines for unsupervised spoken language modeling},
  author={Nguyen, Tu Anh and de Seyssel, Maureen and Roz{\'e}, Patricia and Rivi{\`e}re, Morgane and Kharitonov, Evgeny and Baevski, Alexei and Dunbar, Ewan and Dupoux, Emmanuel},
  journal={arXiv preprint arXiv:2011.11588},
  year={2020}
}

@article{tieu2016npi,
  title={NPI licensing and beyond: Children’s knowledge of the semantics of any},
  author={Tieu, Lyn and Lidz, Jeffrey},
  journal={Language Acquisition},
  volume={23},
  number={4},
  pages={311--332},
  year={2016},
  publisher={Taylor \& Francis}
}

@misc{goriely2024babble,
  author       = {Z{\'e}bulon Goriely and Richard Diehl Martinez and Andrew Caines and Lisa Beinborn and Paula Buttery},
  title        = {From Babble to Words: Pre-Training Language Models on Continuous Streams of Phonemes},
  year         = {2024},
  eprint       = {2410.22906},
  archivePrefix = {arXiv},
  primaryClass = {cs.CL},
  url          = {https://arxiv.org/abs/2410.22906},
}

@inproceedings{warstadt2023babylm,
  author    = {Alex Warstadt and Aaron Mueller and Leshem Choshen and Ethan Wilcox and Chengxu Zhuang and Juan Ciro and Rafael Mosquera and Bhargavi Paranjabe and Adina Williams and Tal Linzen and Ryan Cotterell},
  title     = {Findings of the {BabyLM} Challenge: Sample-Efficient Pretraining on Developmentally Plausible Corpora},
  booktitle = {Proceedings of the BabyLM Challenge at the 27th Conference on Computational Natural Language Learning},
  pages     = {1--34},
  year      = {2023},
  publisher = {Association for Computational Linguistics},
  doi       = {10.18653/v1/2023.conll-babylm.1},
  url       = {https://aclanthology.org/2023.conll-babylm.1},
}

@article{daland2011diphone,
  author    = {Robert Daland and Janet B. Pierrehumbert},
  title     = {Learning Diphone-Based Segmentation},
  journal   = {Cognitive Science},
  volume    = {35},
  number    = {1},
  pages     = {119--155},
  year      = {2011},
  doi       = {10.1111/j.1551-6709.2010.01160.x},
  publisher = {Wiley},
}

@inbook{needle2022phonotactic,
  author    = {Jeremy M. Needle and Janet B. Pierrehumbert and Jennifer B. Hay},
  title     = {Phonotactic and Morphological Effects in the Acceptability of Pseudowords},
  booktitle = {Morphological Diversity and Linguistic Cognition},
  editor    = {Andrew D. Sims and Adam Ussishkin and Jeff Parker and Susannah Wray},
  pages     = {79--112},
  chapter   = {4},
  year      = {2022},
  publisher = {Cambridge University Press},
  doi       = {10.1017/9781108807951.005},
  isbn      = {9781108807951},
}

@article{keuleers2010wuggy,
  author    = {Emmanuel Keuleers and Marc Brysbaert},
  title     = {Wuggy: A Multilingual Pseudoword Generator},
  journal   = {Behavior Research Methods},
  volume    = {42},
  number    = {3},
  pages     = {627--633},
  year      = {2010},
  doi       = {10.3758/BRM.42.3.627},
  publisher = {Springer},
}

@article{saffran1996statistical,
  author    = {Jenny R. Saffran and Richard N. Aslin and Elissa L. Newport},
  title     = {Statistical Learning by 8-Month-Old Infants},
  journal   = {Science},
  volume    = {274},
  number    = {5294},
  pages     = {1926--1928},
  year      = {1996},
  doi       = {10.1126/science.274.5294.1926},
  publisher = {American Association for the Advancement of Science},
}

@book{jusczyk1997discovery,
  author    = {Peter W. Jusczyk},
  title     = {The Discovery of Spoken Language},
  year      = {1997},
  publisher = {MIT Press},
  address   = {Cambridge, MA},
  isbn      = {9780262100489},
}

@article{christophe2008bootstrapping,
  author    = {Anne Christophe and Sara Millotte and S\'{e}verine Bernal
               and Jeffrey Lidz},
  title     = {Bootstrapping Lexical and Syntactic Acquisition Using
               Prosodic and Phonological Constraints},
  journal   = {Language and Speech},
  volume    = {51},
  number    = {1--2},
  pages     = {61--75},
  year      = {2008},
  doi       = {10.1177/00238309080510010501},
  publisher = {SAGE Publications},
}

@article{soderstrom2003prosodic,
  author    = {Melanie Soderstrom and Amanda Seidl and
               Dana Gerber Kemler Nelson and Peter W. Jusczyk},
  title     = {The Prosodic Bootstrapping of Phrases: Evidence from
               Prelinguistic Infants},
  journal   = {Journal of Memory and Language},
  volume    = {49},
  number    = {2},
  pages     = {249--267},
  year      = {2003},
  doi       = {10.1016/S0749-596X(03)00024-X},
  publisher = {Elsevier},
}

@inproceedings{devlin2019bert,
  author    = {Jacob Devlin and Ming-Wei Chang and Kenton Lee
               and Kristina Toutanova},
  title     = {{BERT}: Pre-training of Deep Bidirectional Transformers
               for Language Understanding},
  booktitle = {Proceedings of the 2019 Conference of the North American
               Chapter of the Association for Computational Linguistics:
               Human Language Technologies (Volume 1: Long and Short Papers)},
  pages     = {4171--4186},
  year      = {2019},
  publisher = {Association for Computational Linguistics},
  doi       = {10.18653/v1/N19-1423},
  url       = {https://aclanthology.org/N19-1423},
}

@inproceedings{sennrich2016bpe,
  author    = {Rico Sennrich and Barry Haddow and Alexandra Birch},
  title     = {Neural Machine Translation of Rare Words with Subword Units},
  booktitle = {Proceedings of the 54th Annual Meeting of the Association
               for Computational Linguistics (Volume 1: Long Papers)},
  pages     = {1715--1725},
  year      = {2016},
  publisher = {Association for Computational Linguistics},
  doi       = {10.18653/v1/P16-1162},
  url       = {https://aclanthology.org/P16-1162},
}

@article{warstadt2020blimp,
  author    = {Alex Warstadt and Alicia Parrish and Haokun Liu and
               Anhad Mohananey and Wei Peng and Sheng-Fu Wang and
               Samuel R. Bowman},
  title     = {{BLiMP}: The Benchmark of Linguistic Minimal Pairs
               for {E}nglish},
  journal   = {Transactions of the Association for Computational
               Linguistics},
  volume    = {8},
  pages     = {377--392},
  year      = {2020},
  doi       = {10.1162/tacl_a_00321},
  publisher = {MIT Press},
}

@inproceedings{wang2018glue,
  author    = {Alex Wang and Amanpreet Singh and Julian Michael and
               Felix Hill and Omer Levy and Samuel R. Bowman},
  title     = {{GLUE}: A Multi-Task Benchmark and Analysis Platform for
               Natural Language Understanding},
  booktitle = {Proceedings of the 2018 EMNLP Workshop BlackboxNLP:
               Analyzing and Interpreting Neural Networks for NLP},
  pages     = {353--355},
  year      = {2018},
  publisher = {Association for Computational Linguistics},
  doi       = {10.18653/v1/W18-5446},
  url       = {https://aclanthology.org/W18-5446},
}

@inproceedings{lavechin2022babyslm,
  author    = {Marvin Lavechin and Maureen de Seyssel and Jules
               L{\'{e}}v{\^{e}}que and Emmanuel Dupoux and Alejandrina
               Cristia},
  title     = {{BabySLM}: Language-Acquisition-Friendly Benchmark of
               Self-Supervised Spoken Language Models},
  booktitle = {Proceedings of Interspeech 2022},
  pages     = {1641--1645},
  year      = {2022},
  publisher = {ISCA},
  doi       = {10.21437/Interspeech.2022-10907},
  url       = {https://www.isca-speech.org/archive/interspeech_2022/lavechin22_interspeech.html},
}

@inproceedings{silverman1992tobi,
  author    = {Silverman, Kim and Beckman, Mary and Pitrelli, John and
               Ostendorf, Mari and Wightman, Colin and Price, Patti and
               Pierrehumbert, Janet and Hirschberg, Julia},
  title     = {{ToBI}: A Standard for Labeling {E}nglish Prosody},
  booktitle = {Proceedings of the International Conference on Spoken
               Language Processing (ICSLP)},
  volume    = {2},
  pages     = {867--870},
  year      = {1992},
  doi = {10.21437/ICSLP.1992-260}
}

@techreport{beckman1997tobi,
  author      = {Beckman, Mary E. and Ayers Elam, Gayle},
  title       = {Guidelines for {ToBI} Labelling, Version 3},
  institution = {Department of Linguistics, The Ohio State University},
  year        = {1997},
  url         = {https://www.ling.ohio-state.edu/research/phonetics/E_ToBI/},
}

@misc{ostendorf1995burnc,
  author    = {Ostendorf, Mari and Price, Patti J. and Shattuck-Hufnagel, Stefanie},
  title     = {The {B}oston {U}niversity Radio News Corpus},
  year      = {1995},
  publisher = {Linguistic Data Consortium},
  address   = {Philadelphia},
  note      = {LDC Catalog No.\ LDC95S5},
  doi = {https://doi.org/10.35111/z7xk-z229}
}

@inproceedings{salazar2020pll,
  author    = {Salazar, Julian and Liang, Davis and Nguyen, Toan Q. and
               Kirchhoff, Katrin},
  title     = {Masked Language Model Scoring},
  booktitle = {Proceedings of the 58th Annual Meeting of the Association
               for Computational Linguistics},
  pages     = {2699--2712},
  year      = {2020},
  doi = {https://doi.org/10.48550/arXiv.1910.14659}
}

@article{witten1991,
  author  = {Witten, Ian H. and Bell, Timothy C.},
  title   = {The Zero-Frequency Problem: Estimating the Probabilities of
             Novel Events in Adaptive Text Compression},
  journal = {IEEE Transactions on Information Theory},
  volume  = {37},
  number  = {4},
  pages   = {1085--1094},
  year    = {1991},
  doi={10.1109/18.87000}
}

@book{fenson1993cdi,
  author    = {Fenson, Larry and Dale, Philip S. and Reznick, J. Steven and
               Thal, Donna and Bates, Elizabeth and Hartung, John P. and
               Pethick, Stephen and Reilly, Judy S.},
  title     = {{MacArthur} Communicative Development Inventories:
               User's Guide and Technical Manual},
  publisher = {Singular Publishing Group},
  address   = {San Diego, CA},
  year      = {1993},
}

@article{hubert1985comparing,
  author    = {Hubert, Lawrence and Arabie, Phipps},
  title     = {Comparing Partitions},
  journal   = {Journal of Classification},
  volume    = {2},
  number    = {1},
  pages     = {193--218},
  year      = {1985},
  doi       = {10.1007/BF01908075},
}

@article{ward1963hierarchical,
  author    = {Ward, Joe H.},
  title     = {Hierarchical Grouping to Optimize an Objective Function},
  journal   = {Journal of the American Statistical Association},
  volume    = {58},
  number    = {301},
  pages     = {236--244},
  year      = {1963},
  doi       = {10.1080/01621459.1963.10500845},
}

@article{honnibal2020spacy,
  author = {Honnibal, Matthew and Montani, Ines and Van Landeghem, Sofie and Boyd, Adriane},
  doi = {10.5281/zenodo.1212303},
  title = {{spaCy: Industrial-strength Natural Language Processing in Python}},
  year = 2020
}

@misc{espeakng,
  author = {Duddington, Jonathan and Dunn, Reece H.},
  title = {eSpeak NG: Open source speech synthesizer},
  year = {2025},
  publisher = {GitHub},
  journal = {GitHub repository},
  howpublished = {\url{https://github.com/espeak-ng/espeak-ng}}
}

@article{Bernard2021,
  doi = {10.21105/joss.03958},
  url = {https://doi.org/10.21105/joss.03958},
  year = {2021},
  publisher = {The Open Journal},
  volume = {6},
  number = {68},
  pages = {3958},
  author = {Mathieu Bernard and Hadrien Titeux},
  title = {Phonemizer: Text to Phones Transcription for Multiple Languages in Python},
  journal = {Journal of Open Source Software}
}

@incollection{hay2004speech,
  author    = {Hay, Jennifer and Pierrehumbert, Janet B. and Beckman, Mary E.},
  title     = {Speech perception, well-formedness and the statistics of the lexicon},
  booktitle = {Papers in Laboratory Phonology {VI}},
  editor    = {John Local and Richard Ogden and Rosalind Temple},
  publisher = {Cambridge University Press},
  address   = {Cambridge},
  year      = {2004},
}

@article{lin2001thematic,
  author  = {Lin, Emilie L. and Murphy, Gregory L.},
  title   = {Thematic relations in adults' concepts},
  journal = {Journal of Experimental Psychology: General},
  volume  = {130},
  number  = {1},
  pages   = {3--28},
  year    = {2001},
  doi = {10.1037/0096-3445.130.1.3}
}

@article{imai1994young,
  author  = {Imai, Mutsumi and Gentner, Dedre and Uchida, Noburo},
  title   = {Children's theories of word meaning: The role of shape
             similarity in early acquisition},
  journal = {Cognitive Development},
  volume  = {9},
  number  = {1},
  pages   = {45--75},
  year    = {1994},
  doi = {10.1016/0885-2014(94)90019-1}
}

@article{10.1162/coli.2000.26.4.657,
    author = {MacWhinney, Brian},
    title = {The CHILDES Project: Tools for Analyzing Talk (third edition): Volume {I}: Transcription format and programs, Volume {II}: The database},
    journal = {Computational Linguistics},
    volume = {26},
    number = {4},
    pages = {657-657},
    year = {2000},
    month = {12},
    issn = {0891-2017},
    doi = {10.1162/coli.2000.26.4.657},
    url = {https://doi.org/10.1162/coli.2000.26.4.657},
    eprint = {https://direct.mit.edu/coli/article-pdf/26/4/657/1797615/coli.2000.26.4.657.pdf},
}

@online{fridman2024yannlecun,
  author       = {Lex Fridman and Yann {LeCun}},
  title        = {Yann {LeCun}: Meta {A}{I}, Open Source, Limits of {L}{L}{M}s, {A}{G}{I} \& the Future of {A}{I}},
  year         = {2024},
  url          = {https://lexfridman.com/yann-lecun-3-transcript/},
  note         = {Lex Fridman Podcast \#416 transcript, accessed 2026-05-20}
}

@misc{charpentier2025babylmturns3papers,
      title={Baby{L}{M} Turns 3: Call for papers for the 2025 {B}aby{L}{M} workshop}, 
      author={Lucas Charpentier and Leshem Choshen and Ryan Cotterell and Mustafa Omer Gul and Michael Hu and Jaap Jumelet and Tal Linzen and Jing Liu and Aaron Mueller and Candace Ross and Raj Sanjay Shah and Alex Warstadt and Ethan Wilcox and Adina Williams},
      year={2025},
      eprint={2502.10645},
      archivePrefix={arXiv},
      primaryClass={cs.CL},
      url={https://arxiv.org/abs/2502.10645}, 
}

@inproceedings{bunzeck-etal-2025-small,
    title = "Small Language Models Also Work With Small Vocabularies: Probing the Linguistic Abilities of Grapheme- and Phoneme-Based Baby Llamas",
    author = "Bunzeck, Bastian  and
      Duran, Daniel  and
      Schade, Leonie  and
      Zarrie{\ss}, Sina",
    editor = "Rambow, Owen  and
      Wanner, Leo  and
      Apidianaki, Marianna  and
      Al-Khalifa, Hend  and
      Eugenio, Barbara Di  and
      Schockaert, Steven",
    booktitle = "Proceedings of the 31st International Conference on Computational Linguistics",
    month = jan,
    year = "2025",
    address = "Abu Dhabi, UAE",
    publisher = "Association for Computational Linguistics",
    url = "https://aclanthology.org/2025.coling-main.404/",
    pages = "6039--6048"
}

@inproceedings{bunzeck-zarriess-2025-subword,
    title = "Subword models struggle with word learning, but surprisal hides it",
    author = "Bunzeck, Bastian  and
      Zarrie{\ss}, Sina",
    editor = "Che, Wanxiang  and
      Nabende, Joyce  and
      Shutova, Ekaterina  and
      Pilehvar, Mohammad Taher",
    booktitle = "Proceedings of the 63rd Annual Meeting of the Association for Computational Linguistics (Volume 2: Short Papers)",
    month = jul,
    year = "2025",
    address = "Vienna, Austria",
    publisher = "Association for Computational Linguistics",
    url = "https://aclanthology.org/2025.acl-short.24/",
    doi = "10.18653/v1/2025.acl-short.24",
    pages = "286--300",
    ISBN = "979-8-89176-252-7"
}

\begin{appendices}

\section{Model Architecture and Training}
\label{app:model}
\begin{table}[h]

  \centering
  \small
  \setlength{\tabcolsep}{5pt}
  \begin{tabular}{@{}lcc@{}}
    \toprule
    \textbf{Hyperparameter} & \textbf{10M} & \textbf{100M} \\
    \midrule
    \multicolumn{3}{@{}l}{\textit{Architecture}} \\
    Variant            & BERT-small & BERT-base \\
    Hidden size        & 256        & 768 \\
    Layers             & 8          & 12 \\
    Attention heads    & 8          & 12 \\
    FFN size           & 1{,}024    & 3{,}072 \\
    Vocabulary         & 4{,}000    & 8{,}000 \\
    Parameters         & ${\approx}7.7$M & ${\approx}92$M \\
    \midrule
    \multicolumn{3}{@{}l}{\textit{Training}} \\
    Epochs             & \multicolumn{2}{c}{10} \\
    Max sequence length & \multicolumn{2}{c}{256} \\
    Batch size / device & 1{,}024   & 128 \\
    Gradient accumulation & 1       & 4 \\
    Peak learning rate & $5{\times}10^{-4}$ & $1{\times}10^{-4}$ \\
    LR schedule        & \multicolumn{2}{c}{cosine, 10\% warmup} \\
    Optimizer          & \multicolumn{2}{c}{AdamW} \\
    Weight decay       & \multicolumn{2}{c}{0.01} \\
    Gradient clipping  & \multicolumn{2}{c}{1.0} \\
    Precision          & \multicolumn{2}{c}{bfloat16} \\
    MLM masking        & \multicolumn{2}{c}{15\%} \\
    Random seed        & \multicolumn{2}{c}{42} \\
    \bottomrule
  \end{tabular}
  \caption{Model architecture and training hyperparameters at the two scales.
    Architecture dimensions differ by scale; training settings are shared
    except batch size, gradient accumulation, and peak learning rate.}
  \label{tab:model-config}
\end{table}

\section{Training Corpus and Prosodic Annotation}
\label{app:annotation}

\paragraph{Corpus composition.}
All conditions are trained on the BabyLM 2025 corpus \citep{charpentier2025babylmturns3papers};
Table~\ref{tab:corpus} gives the composition, sampled at the same proportional
weights at both scales.

\begin{table}[h]
  \centering
  \small
  \setlength{\tabcolsep}{4pt}
  \begin{tabular}{@{}lrrr@{}}
    \toprule
    \textbf{Source} & \textbf{10M} & \textbf{100M} & \textbf{Share} \\
    \midrule
    CHILDES (child-directed)   & 2.8M  & 28.9M & 29\% \\
    Project Gutenberg (prose)  & 2.5M  & 26.4M & 26\% \\
    OpenSubtitles (dialogue)   & 2.0M  & 20.0M & 20\% \\
    Simple Wikipedia           & 1.5M  & 14.7M & 15\% \\
    BNC Spoken (conversation)  & 0.9M  & 7.8M  &  8\% \\
    Switchboard (telephone)    & 0.1M  & 1.3M  &  1\% \\
    \midrule
    \textbf{Total}             & 9.95M & 99.0M & 100\% \\
    \bottomrule
  \end{tabular}
  \caption{BabyLM training corpus composition (approximate word counts).}
  \label{tab:corpus}
\end{table}

\paragraph{ToBI inventory.}
The prosodic tokens are drawn from the Tone and Break Indices framework
\citep{silverman1992tobi, beckman1997tobi}. We use two levels of the break-index
hierarchy: \texttt{[B2]} for an intermediate-phrase boundary (ToBI break index 2)
and \texttt{[B4]} for an intonation-phrase boundary (index 4); indices 0, 1, and
3 are not used. We use two boundary tones: \texttt{[H\%]}, a high (rising) tone
associated with yes/no questions, and \texttt{[L\%]}, a low (falling) tone
associated with declaratives, imperatives, and wh-questions. Any annotation
method must be text-only, deterministic, fully automatic, and transparent, since
no paired audio exists for the BabyLM sources.

\paragraph{Rule-based break annotation (\PBreaks{}).}
Each sentence is parsed by spaCy's \texttt{en\_core\_web\_sm} dependency model
\citep{honnibal2020spacy} (NER and lemmatisation disabled for throughput).
\texttt{[B4]} is inserted after any token that is a sentence boundary or bears
terminal punctuation (\texttt{.}, \texttt{?}, \texttt{!}) and takes precedence
over \texttt{[B2]} at the same position. \texttt{[B2]} is inserted, when
\texttt{[B4]} has not been assigned, under the five syntactic cases in
Table~\ref{tab:break-rules}. For sentences of three or fewer whitespace words,
cases A--C and E are suppressed and only case D and the sentence-final
\texttt{[B4]} apply, reflecting the lower parse reliability and negligible
internal structure of short conversational turns. Because standard phonemisation
concatenates the utterance, tokens are placed by a two-pass scheme: each word is
phonemised separately with a temporary separator, then the boundary tokens are
interleaved at the recorded positions.

\begin{table}[h]
  \centering
  \small
  \begin{tabularx}{\columnwidth}{@{}c>{\raggedright\arraybackslash}X@{}}
    \toprule
    \textbf{Rule} & \textbf{Trigger} \\
    \midrule
    A & Coordinating clause junction: a coordinating conjunction heads a
        clause-level \texttt{cc} dependency on a verbal or auxiliary root. \\
    B & Subordinate or complement clause onset: a subordinating conjunction or
        complementiser marks an adverbial, complement, or embedded clause. \\
    C & Relative clause onset: a relative pronoun or the \texttt{relcl} arc. \\
    D & Minor surface punctuation: comma, semicolon, colon, dash, or ellipsis. \\
    E & Appositive or parenthetical nominal: a token bearing the \texttt{appos}
        dependency and not preceded by punctuation. \\
    \bottomrule
  \end{tabularx}
  \caption{Syntactic cases that trigger a \texttt{[B2]} token in \PBreaks{}.}
  \label{tab:break-rules}
\end{table}

\paragraph{Neural annotator (\ProsodyNN{}).}
\ProsodyNN{} replaces the rule-based break detector with a trained sequence
labeller while keeping the same token vocabulary and two-pass pipeline. The
annotator is trained on the Boston University Radio News Corpus
\citep{ostendorf1995burnc}, the only large manually ToBI-annotated American
English corpus, comprising roughly 3{,}600 sentences and 113{,}000 word tokens
of read radio news after excluding one speaker who lacks break annotation. Break
indices are collapsed to \{NONE, B2, B4\} and tone events (mapped to the nearest
following word boundary) to \{NONE, L\%, H\%\}; splits are speaker-disjoint. The
model is a \texttt{bert-base-uncased} backbone \citep{devlin2019bert} with two
linear heads over the first-subword hidden state of each word, trained to
minimise a joint class-weighted cross-entropy
$\mathcal{L} = \mathcal{L}^{\text{brk}} + \lambda\,\mathcal{L}^{\text{ton}}$
with weights $[1,3,3]$ (break) and $[1,4,8]$ (tone); the elevated \texttt{[H\%]}
weight compensates for the rarity of yes/no questions in read news. Training
uses AdamW at learning rate $2{\times}10^{-5}$ with early stopping on held-out
macro-$F_1$.

Each head is used for corpus annotation only if it clears a held-out $F_1$ gate
(Table~\ref{tab:annotator-f1}); the break head otherwise falls back to the
rule-based \PBreaks{} pipeline and the tone head to the heuristic below. The
final annotator clears every threshold, so both heads are shipped. Inference over
the full 100M corpus takes about an hour on a single GPU.

\begin{table}[h]
  \centering
  \small
  \setlength{\tabcolsep}{6pt}
  \begin{tabular}{@{}lcc@{}}
    \toprule
    \textbf{Head} & \textbf{$F_1$} & \textbf{Ship threshold} \\
    \midrule
    \texttt{[B2]} & 0.58 & $\geq 0.50$ \\
    \texttt{[B4]} & 0.99 & $\geq 0.80$ \\
    \texttt{[L\%]} & 0.69 & $\geq 0.60$ \\
    \texttt{[H\%]} & 0.56 & $\geq 0.30$ \\
    \bottomrule
  \end{tabular}
  \caption{Held-out $F_1$ of the neural annotator against the shipping
    thresholds. All four clear their gate, so the full annotator is used.}
  \label{tab:annotator-f1}
\end{table}

\paragraph{Rule-based tone heuristic (fallback).}
When the tone head fails its gate, sentence-final tone is assigned by a
three-gate rule, which is also the source of the boundary tones' definition.
A sentence whose final non-bracket character is not \texttt{?} receives
\texttt{[L\%]}. Otherwise, the first substantive word is found by skipping up to
five leading discourse markers, fillers, and response tokens: a wh-word receives
\texttt{[L\%]}, a fronted positive auxiliary or copula receives \texttt{[H\%]}
(contracted negations are excluded as they often head echo questions), and any
remaining question receives \texttt{[L\%]} conservatively. In our run the tone
head cleared its gate, so this heuristic was not triggered.

\section{Pseudoword Generation and Phonotactic Control}
\label{app:wuggy-control}
\paragraph{Generation: a proposal stage and a selection stage.}
Pseudowords are produced in two stages. The proposal stage supplies candidate
forms and the selection stage decides which candidates enter the test set. The
proposal stage is a trigram Markov chain over orthographic characters, trained
on the alphabetic types of the training vocabulary, sampled autoregressively
from $\langle s\rangle\langle s\rangle$ until $\langle/s\rangle$ with a
temperature $T$ that reshapes the empirical transition counts as
$P_T \propto c^{1/T}$. Candidates are constrained to between four and seven
characters. Each candidate is then phonemised by the same eSpeak-NG pipeline
that produced the training corpora, and every subsequent step operates on the
resulting IPA form: the collision filters, the phonotactic score $S$, the
length and score matching, and the construction of the Inverted category. The
chain is therefore a proposal distribution only. None of the properties the
test set relies on are inherited from it, and all of them are imposed in
phonemic space. This is what separates the design from a Wuggy-style pipeline,
which controls sub-syllabic substitutions in the orthographic domain and
phonemises afterwards, leaving the phonemic properties of the pair
uncontrolled.
\paragraph{Selection and pairing.}
Real words are drawn from the Filter-A-validated pool
(Appendix~\ref{app:lexical}) by stratified sampling over ten score strata,
restricted to items of four to seven IPA phones. Candidate pseudowords are
generated at five times the required pool size for the standard categories and
screened by Filter B, with rejections backfilled so category sizes are held
fixed. Each real word is then matched to the pseudoword nearest it in
phonotactic score \emph{drawn only from candidates with the same number of IPA
phones}, subject to $|S_{\text{real}} - S_{\text{fake}}| \le 2.0$. Every one of
the $4{,}129$ pairs is exactly length-matched in phones as a result; the mean
absolute score difference is $0.19$ for LexicalMemory and $0.16$ for
Composition. The set comprises $967$ LexicalMemory, $1{,}162$ Composition and
$2{,}000$ Inverted pairs.

\paragraph{Phonotactic control in the BabySLM benchmark:}
The same phonemic control is absent from the benchmark our task replaces. Wuggy controls the biphone statistics of its pseudowords, but not triphone statistics, which are known to carry wordlikeness \citep{bailey2001determinants, hay2004speech, needle2022phonotactic}. Table~\ref{tab:wuggy-control} scores all $17{,}706$ real/pseudoword pairs of the BabySLM lexical development set under both models. Under the biphone model Wuggy does control, the real word is favoured on $58.1\%$ of trials, close to the residual asymmetry expected from real words being better formed on average. Under the uncontrolled triphone model it is favoured on $70.3\%$. A model that had learned nothing beyond triphone phonotactics would therefore score close to 70\% here, despite having no lexical knowledge at all.

\begin{table}[h]
  \centering
  \small
  \setlength{\tabcolsep}{6pt}
  \begin{tabular}{@{}lcc@{}}
    \toprule
    & \textbf{Biphone} & \textbf{Triphone} \\
    & (controlled) & (uncontrolled) \\
    \midrule
    Real word favoured   & 58.1\%      & 70.3\% \\
    Mean difference      & $+0.31$     & $+1.21$ \\
    Median difference    & $+0.21$     & $+0.89$ \\
    Skew                 & $+0.56$     & $+0.75$ \\
    Mean $-$ median      & $+0.10$     & $+0.32$ \\
    Cohen's $d$          & 0.13        & 0.28 \\
    KS $D$               & 0.070       & 0.128 \\
    Wilcoxon             & $p{=}1.5{\times}10^{-157}$ & $p{<}10^{-300}$ \\
    \bottomrule
  \end{tabular}
  \caption{Phonotactic asymmetry between the real words and the Wuggy
    pseudowords of the BabySLM lexical development set ($n = 17{,}706$ pairs),
    under a biphone model, which Wuggy controls, and a triphone model, which it
    does not.}
  \label{tab:wuggy-control}
\end{table}

\section{Lexical Decision: Filtering and Examples}
\label{app:lexical}

\paragraph{The two filters.}
\emph{Filter A} (real-word validity) admits a candidate real word only if it is
attested by \texttt{wordfreq} \emph{and}, after WordNet lemmatisation, present in
the NLTK \texttt{words} list. \texttt{wordfreq} alone admits web artefacts such
as \textit{https} and repeated-character strings, while the NLTK list alone lacks
inflected forms; gating one on the other keeps inflected and loan words while
removing non-words. \emph{Filter B} (pseudoword screen) rejects a generated
pseudoword if its orthographic form occurs in the training vocabulary, or if its
eSpeak IPA form, either verbatim or with stress and length marks stripped,
occurs in the real-word IPA lexicon (the set of IPA forms of all corpus words).
Rejected candidates are resampled so category counts are held fixed.

\paragraph{Contamination.}
Filter A reduced the real-word pool from the $82{,}548$ alphabetic corpus types
to $43{,}947$ validated English words, rejecting $38{,}601$ ($47\%$) as
non-words. Filter B then removed a substantial number of pseudowords whose
phonemic form coincided with a real word (Table~\ref{tab:contamination}); in the
Inverted category the $3{,}244$ IPA collisions exceed the $2{,}000$ pairs finally
retained, confirming that without the phonemic screen a large share of
``pseudowords'' would in fact be real words in IPA space.

\begin{table}[h]
  \centering
  \small
  \setlength{\tabcolsep}{6pt}
  \begin{tabular}{@{}lc@{}}
    \toprule
    \textbf{Category} & \textbf{Filter B IPA collisions removed} \\
    \midrule
    LexicalMemory & 959 \\
    Composition   & 1{,}053 \\
    Inverted      & 3{,}244 \\
    \bottomrule
  \end{tabular}
  \caption{Pseudowords rejected by the phonemic screen (Filter B) because their
    IPA form matched a real corpus word.}
  \label{tab:contamination}
\end{table}

\paragraph{Example pairs.}
Table~\ref{tab:pair-examples} shows representative pairs from each category. In
LexicalMemory and Composition the phonotactic scores are matched within the
$2.0$ threshold; in Inverted the pseudoword's score is far higher than the real
word's by construction.

\begin{table}[h]
  \centering
  \small
  \setlength{\tabcolsep}{4pt}
  \begin{tabular}{@{}llrr@{}}
    \toprule
    \textbf{Real} & \textbf{Pseudoword} & $\mathbf{S_{\text{real}}}$ & $\mathbf{S_{\text{fake}}}$ \\
    \midrule
    \multicolumn{4}{@{}l}{\textit{LexicalMemory}} \\
    pencil \ipa{/pɛnsəl/}  & blaver \ipa{/bleɪvɚ/} & $-18.7$ & $-18.7$ \\
    heart \ipa{/hɑːɹt/}    & pred \ipa{/pɹɛd/}     & $-11.9$ & $-12.0$ \\
    \multicolumn{4}{@{}l}{\textit{Composition}} \\
    reissued \ipa{/ɹiːɪʃuːd/} & pophini \ipa{/pəfiːni/} & $-29.1$ & $-29.1$ \\
    foppish \ipa{/fɑːpɪʃ/}    & cenfse \ipa{/sɛnfs/}    & $-25.0$ & $-25.0$ \\
    \multicolumn{4}{@{}l}{\textit{Inverted}} \\
    gestalt \ipa{/ɡəʃtɑːlt/} & plation \ipa{/pleɪʃən/} & $-39.9$ & $-13.2$ \\
    kinetic \ipa{/kᵻnɛɾɪk/}  & noment \ipa{/noʊmənt/}  & $-37.5$ & $-13.3$ \\
    \bottomrule
  \end{tabular}
  \caption{Example real-word / pseudoword pairs with Witten-Bell triphone
    phonotactic scores $S$. Standard pairs are score-matched; Inverted pairs
    satisfy $S_{\text{fake}} > S_{\text{real}}$.}
  \label{tab:pair-examples}
\end{table}

\section{Target Vocabulary}
\label{app:semantic}

The semantic tasks use the 68 CDI words \citep{fenson1993cdi} listed in
Table~\ref{tab:cdi-vocab}, drawn from the earliest vocabulary of
English-speaking children and high-frequency in the CHILDES development set of the
BabyLM corpus. The six categories are concrete semantic fields whose categorical
structure is well established developmentally, so the benchmark tests whether the
distributional structure the model has seen is reflected in its embedding
geometry rather than generalisation to novel words.

\begin{table}[h]
  \centering
  \small
  \begin{tabularx}{\columnwidth}{@{}l>{\raggedright\arraybackslash}X@{}}
    \toprule
    \textbf{Category} & \textbf{Words} \\
    \midrule
    Animal    & dog, cat, bird, fish, bear, duck, horse, cow, pig, frog,
                mouse, sheep \\
    Food      & milk, juice, water, cookie, apple, banana, bread, cheese, cake,
                egg, soup, rice \\
    Body part & eye, nose, mouth, ear, hand, foot, head, hair, finger, knee,
                arm, leg \\
    Clothing  & shoe, sock, hat, shirt, coat, dress, boot, sweater, diaper,
                button, pocket \\
    Vehicle   & car, bus, truck, bike, boat, plane, wagon, stroller, tractor \\
    Household & cup, spoon, bowl, plate, bottle, chair, table, bed, blanket,
                door, phone, clock \\
    \bottomrule
  \end{tabularx}
  \caption{The 68 CDI target words by semantic category \citep{fenson1993cdi}.}
  \label{tab:cdi-vocab}
\end{table}

\section{Tokenisation}
\label{app:tokenisation}

\paragraph{Training.}
Every condition trains its own Byte-Pair Encoding tokeniser
\citep{sennrich2016bpe} from scratch on its own preprocessed corpus, using the
HuggingFace \texttt{tokenizers} implementation. Vocabulary size is 4{,}000 at
10M and 8{,}000 at 100M, chosen to hold the ratio of vocabulary to model
parameters roughly constant across the two scales; the minimum merge frequency
is left at the library default of 2, and no merge-count or vocabulary sweep was
run. The 10M orthographic tokeniser learns 3{,}708 merges and the 10M
\PGoriely{} tokeniser 3{,}938, the balance of each vocabulary being the base
character inventory. Out-of-vocabulary material falls back to \texttt{[UNK]},
and BPE dropout is not used.

Structural tokens (\texttt{[B2]}, \texttt{[B4]}, \texttt{[H\%]},
\texttt{[L\%]}, \texttt{[UTT]}, \texttt{[SPEAKER]},
\texttt{[HEADER]}, and the reserved BERT tokens) are filtered out of the corpus
before BPE training and reserved at fixed vocabulary ids by the trainer. Without
this the pre-tokeniser splits \texttt{[SPEAKER]} into \texttt{[}, \texttt{SPEAKER},
\texttt{]} and BPE learns merges inside the token name.

\paragraph{Segmentation.}
At application time BPE begins from the character sequence and repeatedly
applies the highest-priority applicable merge from the learned merge list, in
the order the merges were learned. It is not a longest-match scan: a longer
entry in the vocabulary will not be preferred if the merges that would build it
rank below the merges that build a different segmentation.

\paragraph{Pre-tokenisation, and the asymmetry between conditions.}
All conditions use the same whitespace pre-tokeniser, but it does very different
work in each. \Ortho{} retains ordinary inter-word whitespace, so the
pre-tokeniser splits the stream at every word boundary and no merge can span
one: word boundaries are preserved by construction, and the model receives them
for free. The phonemic conditions have inter-word whitespace removed, which is
the defining property of the \citet{goriely2024babble} setting we build on, so
the pre-tokeniser splits only around structural tokens and effectively presents
BPE with the whole utterance as a single unit. Merges may therefore span word
boundaries, and no word boundary is marked in the input at any point.

This asymmetry is deliberate rather than incidental: it is what modelling
continuous speech without orthographic segmentation means. It does, however,
mean that the orthographic baseline is handed a source of information the
phonemic conditions must infer.

\section{Token Budget by Condition}
\label{app:token-budget}

Annotating a phoneme stream with structural tokens necessarily lengthens it,
which raises the question of whether any advantage of the annotated conditions
is a token-budget effect rather than a representational one. Each unified
corpus was tokenised with its own trained tokeniser, with
\texttt{add\_special\_tokens=False} so that no \texttt{[CLS]} or \texttt{[SEP]}
padding is counted.

\begin{table*}[t]
  \centering
  \footnotesize
  \setlength{\tabcolsep}{7pt}
  \begin{tabular}{@{}lrrrrrr@{}}
    \toprule
    \textbf{Condition} & \textbf{Tokens} & \textbf{Per utt.}
      & \textbf{Structural} & \textbf{\% struct.}
      & \textbf{vs \PGoriely} & \textbf{vs \Ortho} \\
    \midrule
    \multicolumn{7}{@{}l}{\textit{10M}} \\
    \Ortho      & 14{,}979{,}765 & 12.85 & 0             & 0.00\%  & $+22.0\%$ & n/a \\
    \PGoriely   & 12{,}276{,}352 & 10.55 & 1{,}010{,}527 & 8.23\%  & n/a       & $-18.0\%$ \\
    \PBreaks    & 13{,}498{,}252 & 11.59 & 2{,}152{,}618 & 15.95\% & $+10.0\%$ & $-9.9\%$ \\
    \ProsodyNN  & 13{,}417{,}556 & 11.52 & 2{,}104{,}219 & 15.68\% & $+9.3\%$  & $-10.4\%$ \\
    \midrule
    \multicolumn{7}{@{}l}{\textit{100M}} \\
    \Ortho      & 140{,}210{,}835 & 12.24 & 0              & 0.00\%  & $+27.8\%$ & n/a \\
    \PGoriely   & 109{,}677{,}972 & 9.58  & 9{,}875{,}227  & 9.00\%  & n/a       & $-21.8\%$ \\
    \PBreaks    & 122{,}184{,}461 & 10.67 & 21{,}295{,}673 & 17.43\% & $+11.4\%$ & $-12.9\%$ \\
    \ProsodyNN  & 121{,}277{,}565 & 10.59 & 20{,}780{,}910 & 17.13\% & $+10.6\%$ & $-13.5\%$ \\
    \bottomrule
  \end{tabular}
  \caption{BPE token counts per condition and scale, with the share contributed
    by structural tokens and the budget relative to \PGoriely{} and to
    \Ortho{}.}
  \label{tab:token-budget}
\end{table*}

\paragraph{Composition of the structural share.}
At 10M, \PBreaks{} spends 795{,}812 tokens on \texttt{[B2]} and 1{,}356{,}806
on \texttt{[B4]}, and \ProsodyNN{} spends 883{,}401 on \texttt{[B2]},
1{,}166{,}168 on \texttt{[B4]}, 17{,}723 on \texttt{[H\%]} and 36{,}927 on
\texttt{[L\%]}. At 100M the corresponding counts are 7{,}852{,}083 and
13{,}443{,}590 for \PBreaks{}, and 8{,}757{,}921, 11{,}465{,}427, 172{,}011 and
385{,}551 for \ProsodyNN{}. \PGoriely{} carries only \texttt{[UTT]}, 1{,}010{,}527 at 10M and
9{,}875{,}227 at 100M. Neither annotated condition emits it: in both, the
utterance-final position is carried by \texttt{[B4]}. The two annotators differ
in how often they place \texttt{[B4]} utterance-internally, the rule-based
annotator doing so on roughly 192{,}000 lines at 10M against roughly 1{,}700
for the neural annotator, which is why \PBreaks{} carries slightly more
structural tokens than \ProsodyNN{} despite emitting no tones.

\paragraph{Interpretation.}
Prosodic annotation does inflate the budget within the phonemic family, by
9.3\% to 10.0\% at 10M and 10.6\% to 11.4\% at 100M relative to \PGoriely{}, so
the confound is real for comparisons between the annotated conditions and
\PGoriely{}. It cannot, however, explain any phonemic advantage over
orthography, because every phonemic condition still consumes markedly fewer
tokens than \Ortho{}: the annotated conditions sit 9.9\% to 10.4\% below at 10M
and 12.9\% to 13.5\% below at 100M, and \PGoriely{} is 18.0\% to 21.8\% below.
The orthographic baseline has the largest token budget of all eight models and
is nonetheless overtaken at 100M on unsupervised category induction. The
asymmetry runs against the annotated conditions, not in their favour.

%%Always put this at the end to avoid formatting issues

\begin{figure*}[t]
  \section{UMAP Projections - Semantic Category Structure}
  \label{app:umap}
  \centering
  \includegraphics[width=0.9\textwidth]{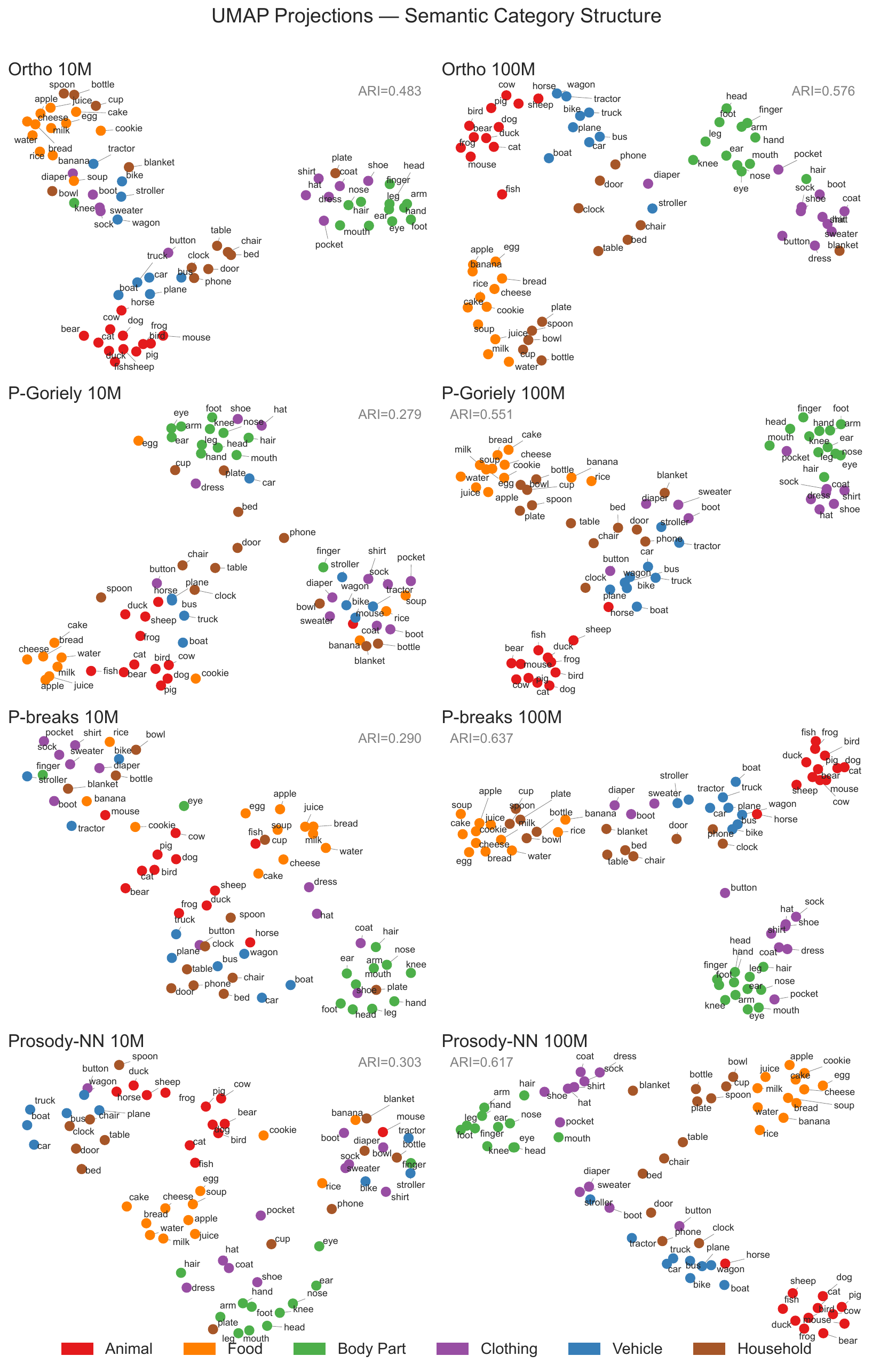}
  \caption{UMAP projections of the 68 CDI word embeddings, coloured by semantic
    category, for \Ortho{} and \ProsodyNN{} at both scales; $k$-means ARI is
    annotated in each panel.}
  \label{fig:umap}
\end{figure*}

\end{appendices}
\end{document}